\documentclass[conference]{IEEEtran}
\IEEEoverridecommandlockouts
 \pdfoutput=1
\usepackage{amsfonts}
\usepackage{algorithmic}
\usepackage{graphicx}
\usepackage{textcomp}
\usepackage{xcolor}
\usepackage{comment}
\usepackage{hyperref}
\usepackage{rotating}
\usepackage{color, colortbl}
\usepackage{multicol}
\usepackage{multirow}
\usepackage[center]{caption}
\usepackage{graphics} 
\usepackage{epsfig} 
\usepackage{times} 
\usepackage{amsmath} 
\usepackage{amssymb} 
\usepackage{cite}
\usepackage{url}
\usepackage{dblfloatfix}
\usepackage{float}

\def\BibTeX{{\rm B\kern-.05em{\sc i\kern-.025em b}\kern-.08em
    T\kern-.1667em\lower.7ex\hbox{E}\kern-.125emX}}
\begin{document}
\begin{onecolumn}
\Huge{\textbf{IEEE Copyright Notice}}

\vspace{5em}

\large{\textcopyright 2021 IEEE. Personal use of this material is permitted. Permission from IEEE must be obtained for all other uses, in any current or future media, including reprinting/republishing this material for advertising or promotional purposes, creating new collective works, for resale or redistribution to servers or lists, or reuse of any copyrighted component of this work in other works.}

\vspace{10em}

\Large{Accepted to be published in: IEEE Aerospace and Electronic Systems Magazine}

\end{onecolumn}

\title{Search Planning of a UAV/UGV Team with Localization Uncertainty in a Subterranean Environment}
\author{\IEEEauthorblockN{ Matteo De Petrillo\IEEEauthorrefmark{2}, Jared Beard, Yu Gu, Jason N. Gross}

\thanks{\textit{Authors are with the Department of Mechanical and Aerospace Engineering, } 
\textit{West Virginia University}, Morgantown, USA. 
}
 \thanks{This research is funded by an academic grant from the National Geospatial-Intelligence Agency (Award No. HM0476-18-1-2000, Project Title: ``Exploration Planning of a UAV/UGV Team with Localization Uncertainty in a Subterranean Environment''). Approved for public release, 21-286.}
\thanks{
\IEEEauthorrefmark{2}{madepetrillo@mix.wvu.edu }}
}
\twocolumn
\maketitle
\thispagestyle{empty}
\pagestyle{empty}

\begin{abstract}
We present a waypoint planning algorithm for an unmanned aerial vehicle (UAV) that is teamed with an unmanned ground vehicle (UGV) for the task of search and rescue in a subterranean environment. The UAV and UGV are teamed such that the localization of the UAV is conducted on the UGV via the multi-sensor fusion of a fish-eye camera, 3D LIDAR, ranging radio, and a laser altimeter. Likewise, the trajectory planning of the UAV is conducted on the UGV, which is assumed to have a 3D map of the environment (e.g., from Simultaneous Localization and Mapping). The goal of the planning algorithm is to satisfy the mission’s exploration criteria while reducing the localization error of the UAV by evaluating the belief space for potential exploration routes. The presented algorithm is evaluated in a relevant simulation environment where the planning algorithm is shown to be effective at reducing the localization errors of the UAV.
\end{abstract}

\section{Introduction}
 Nowadays, the use of robots in everyday life is becoming more usual for a variety of scenarios. Among the most commonly proposed applications include: maintenance services \cite{katrasnik2009survey} \cite{preston2017use}, industrial facility inspection \cite{nikolic2013uav}, and search and rescue \cite{sampedro2019fully}. For search and rescue application scenarios in hazardous environments, there is a clear need to replace some of the first responders with mobile robots. In this scenario, mobile robots, not only can save lives but could also perform better due to the variety of sensors available and the fact that they are less susceptible to conditions like dust or poor lighting conditions. However, many challenges must be addressed in order to have a fully autonomous robot or team of robots in these environments. Of the many challenges, an important difficulty arises from attempting to navigate in these environments due to lack of access to Global Navigation Satellite Systems (GNSS) signals, and also unknown obstacles. 

 Extensive research has been done in the use of robots in subterranean environments as is discussed in \cite{morris2006recent}. With a specific focus on underground mines, Thrun S. at al. \cite{thrun2004autonomous} presented a ground rover designed to explore and map these harsh environments. Nowadays, technology and the miniaturization of electronic components has allowed for the reduction of the size of UAVs, such as commercial drones, making it possible to leverage their ability to traverse a variety of environments and complete a large number of tasks. For example, Azhari F. et al. \cite{azhari2017comparison} used a UAV for underground mine mapping, taking advantage of its six degrees of freedom. To autonomously navigate in subterranean tunnels or mines, Papachristos C. et al. \cite{papachristos2019autonomous} provided an unmanned aerial system capable of performing simultaneous localization and mapping (SLAM) fusing different sensors to provide valuable information in dark conditions, presence of dust, etc. The authors also provided an uncertainty-aware path planning strategy for autonomous micro UAVs to balance exploration and mapping in real-time \cite{papachristos2019localization}.

 Due to the limited capability of a single UAV to carry a considerable amount of sensors, the use of multiple robots to achieve a common task is also well studied in the literature. One of the many benefits of using multiple robots consists of reducing the time to perform the task, such as exploration. In these cases, the planning algorithm is optimized to reduce the time to collect information \cite{simmons2000coordination} choosing different waypoints \cite{burgard2000collaborative}, especially if the same kind of robot and sensors are used. For example, to diversify information and to perform both autonomous search and rescue, we presented in previous work a UAV and UGV team for subterranean exploration \cite{gross2019field}. The cooperation of multiple robots increases the ability to perform a wider variety of tasks. Li J. et al. \cite{7728117} use a UAV to provide a higher point of view to improve the environment map for the UGV navigation. 
 
 When teaming robots for exploration, it is most common to assume that each agent is capable of estimating their pose, such that the team is primarily used to improve the efficiency of environment mapping and exploration. This is the approach that was adopted in\cite{qin2019}. Similarly, Delmerico J. et al. \cite{delmerico2017} used a UAV, capable of self-localization, to improve and expand the map exploration for a ground robot and to optimize the mission time. However, having the UAVs in the team maintaining accurate self-localization presents challenges. For example, as discussed in Baldini F. et al. \cite{baldini2019}, using Visual-Inertial Odometry (VIO) solutions on UAVs is prone to solution drift, and the level of drift is heavily affected by the light condition in which the UAV operates. For these reasons, \cite{baldini2019} used a learning approach to reduce drift. To reduce drift, employing Simultaneous Localization and Mapping (SLAM) is possible directly on a UAV as has been demonstrated in\cite{valenti2014}. However, some operating environments (e.g., a long corridor or tunnel), may affect the reliability of loop-closures, crucial for SLAM.

 To address some of these challenges, the approach presented in this work proposes to use active sensors on the UGV to localize the UAV with respect to the UGV in a manner that is not prone to drift, and with modalities that will work in a variety of lighting conditions. In particular, following on the approach considered in our initial work presented in \cite{gross2019field}. With this UAV/UGV teaming configuration, the UGV has the advantage of additional load capacity and longer endurance to easily carry many sensors along with the significant computational power needed to use them in near real-time. In this scenario, the UGV is used to map the environment as well as for estimating the state of the UAV within the map, while the UAV's mobility is leveraged to perform the search task. However, when adopting this approach, due to the nature of the UGV's sensors' uncertainty, as well as the potential for non over-lapping sensor fields-of-view, the localization uncertainty of the UAV is affected by the chosen flight path. As such, the focus of this paper is to develop a path planning algorithm for the UAV that takes into consideration both the exploration needs and the localization uncertainty of the UAV. 
 
 Considering the position uncertainty in the planning algorithm consists of finding a configuration in the belief space such that motion increases information from sensing, and reduces pose uncertainty of the robot with respect to the environment \cite{patil2015scaling}. Planning under uncertainty leads to the use of a dual-layer architecture; where the first layer predicts all the possible outcomes and the second layer determines the best action to take \cite{indelman2015planning}. For mobile robot path planning, Prentice S. et al. \cite{prentice2009belief} presented a revised Probabilistic Road Map (PRM) \cite{kavraki1998analysis}, called Belief RoadMap, where they fuse the predicted position estimated uncertainty into the planning process. In this work, the trajectory was designed with respect to the location of pre-deployed ranging radio beacons in order to reduce the position uncertainty while traversing from a start to a goal location. A belief space planning problem is often formulated as a partially observable Markov decision process (POMDP) and Van De Berg J. et al. \cite{van2012motion}, similar to the approach presented in this paper, used an extended Kalman filter to approximate the robot belief dynamics.
 
 Different from prior belief space planning research, our approach has a required exploration goal and seeks to accomplish it such that the UAV's localization uncertainty is maintained within an acceptable level during the operation. This paper contributes a new search planning algorithm that fuses the capabilities of two robots (UAV/UGV) to balance between environment exploration tasks while reducing the localization error for the UAV along the selected trajectory. This paper substantively builds upon our previous work \cite{gross2019field}, where the initial design and evaluation of the UAV localization system is presented. 
 
 The rest of this paper is organized into five sections. Section II provides a description of the system where the assumptions and constraints of the problem are presented; section III explains the path planning algorithm; section IV presents the simulation environment; section V discusses the results and section VI offers conclusions and future work.
 
\section{System description}
\subsection{Assumptions and Constraints}\label{ass_cons}
 In this work, the UAV and UGV systems are modeled according to the developed physical hardware system that is described in \cite{gross2019field}.
 \begin{figure}[H]
     \centering
        \includegraphics[width=.5\linewidth]{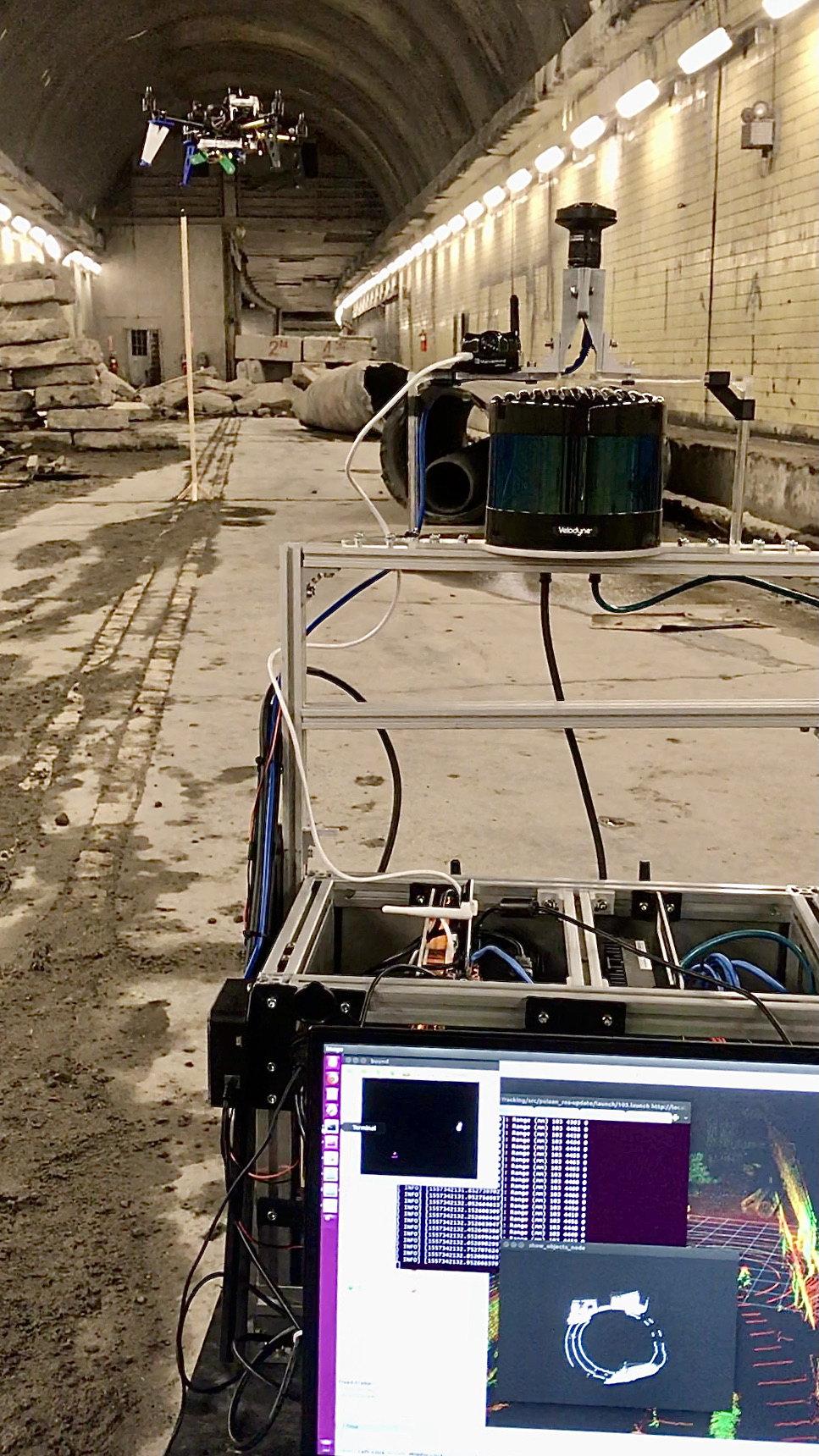}
          \caption{Underground highway tunnel for testing purposes. In the foreground the UGV equipped with the main localization sensors and in the background the UAV flies in front of the LIDAR}
     \label{fig:realT}
 \end{figure}
 
 As shown in Figure \ref{fig:realT}, a subterranean environment would block a possible positioning system, such as GNSS, and for this reason different sensors need to be used for localization purposes. 
 The UAV is assumed to carry a downward-facing camera to perform the search, a LIDARLite altimeter that provides readings at 5 Hz, and an Inertial Measurement Unit (IMU) with updates at 50 Hz. The UGV is assumed to be equipped with a 128 channel 3D LIDAR and a fish-eye camera that is pointed upward and both of which are assumed to provide measurements at 10 Hz update rate. To increase the LIDAR field-of-view for the UAV detection, the LIDAR is pitched 15 degrees upward. The fish-eye camera is mounted at 0.8 m above the ground such that its field of view is restricted to everything above that height. Finally, both the UAV and UGV have a pair of Ultra-Wideband (UWB) ranging radios to provide the relative range measurement between the two vehicles at 10 Hz. 
 
\begin{figure}[!h]
    \centering
     \includegraphics[width=\columnwidth]{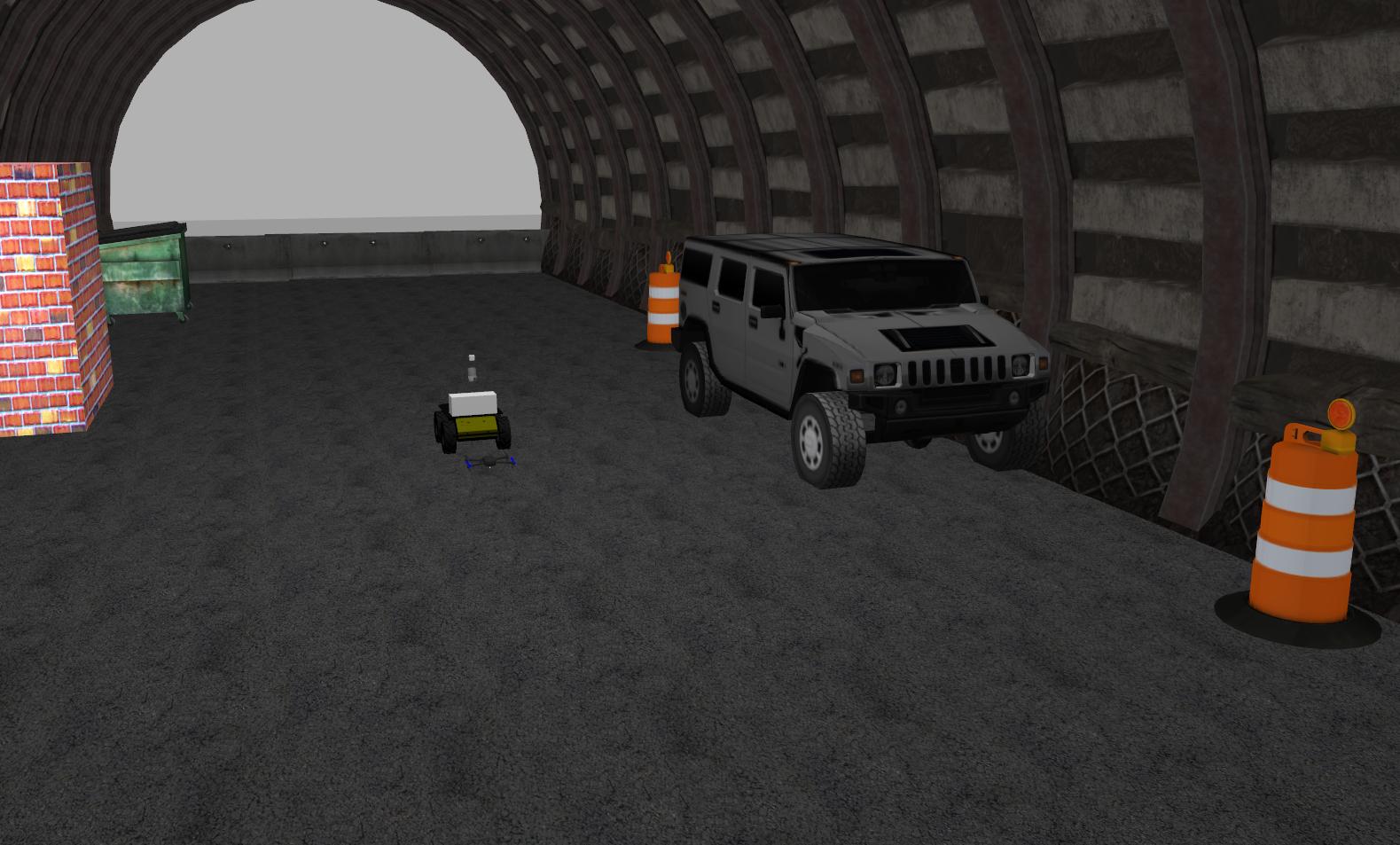}
    \caption{Gazebo simulation environment that reflects a highway tunnel with some obstacles}
    \label{tunnel1}
\end{figure}

 The simulation for this work is modeled after a real-world test environment, shown in Figure \ref{fig:realT}, that consists of an underground highway tunnel section which is 40 meters long, 10 meters wide, and 8 meters high with various obstacles as shown in Figure \ref{tunnel1}. 
 
 In this work, it is assumed that the UAV is deployed from the UGV and returned to the UGV after a search, therefore, the planning algorithm is designed to provide a path that starts and ends at the UGV's location after exploring the entire map. Further, while the UAV performs its search mission, the UGV is assumed to be static to provide the UAV's localization estimate. This constraint was chosen in order to simplify the relative localization between the UAV and UGV.

Between each waypoint, to support the belief state propagation during the planning, the sensor updates are simulated by assuming the UAV moves with zero acceleration at a constant velocity $v = 0.5\text{ m/s}$ with pitch and roll of zero. Further, the UAV is assumed to fly along straight lines connecting the waypoints. These assumptions were chosen to simplify the process of determining observation models and the number of sensor updates given the assumed update rates when propagating the belief state. Further, during the belief state propagation process the camera update was assumed to occur whenever the UAV was within $6 \textit{ meters}$ range. This threshold was introduced to consider the fish-eye distortion and resolution and was determined via testing the camera tracking performances. 

A constraint taken into account by the planning algorithm is the number of waypoints visited, such that the UAV's flight time is limited 
\begin{equation}
    T_{flight} < \rho
\end{equation}
where $\rho$ is a specific threshold in seconds determined by the Quad-rotor characteristics and payload. 

The UAV flight controller is set to follow the waypoint trajectory provided by the planner, and is formulated to balance exploration and at the same time to keep the localization uncertainty under a reasonable value $\delta$
\\
\begin{equation}
   \left\Vert
      P_{pos} 
      \right\Vert_2
     \label{delta} < \delta
\end{equation}
with $P_{pos}$ being the covariance matrix P relative to the UAV's position. Since this work focuses on the ability of the planning algorithm to differentiate between pre-selected paths, $\delta$, in Eq. \ref{delta}, needs to be a number small enough to guarantee that the UAV does not collide with any obstacle. This value depends on the environment and the density of the obstacles present in it.

The sensors' measurements, used to estimate the UAV position, are assumed to always be available from the altimeter and Ultra-Wideband (UWB) ranging radio, whereas for the fish-eye camera and the LIDAR measurement updates occur when the UAV is in their respective fields of view.

\section{Planning Algorithm}
The planning algorithm is required to provide a path for the UAV to follow to perform a search while also meeting the physical constraints. Before planning occurs, it is assumed that the UGV has already produced an acceptable reconstruction of the environment using simultaneous localization and mapping (SLAM). The planner, which is assumed to be performed before the flight and offline, uses the 3D map provided by the UGV as an input. 
\subsection{Graph Construction Phase}
\begin{figure}[!h]
    \centering
 \includegraphics[width = 0.8\columnwidth]{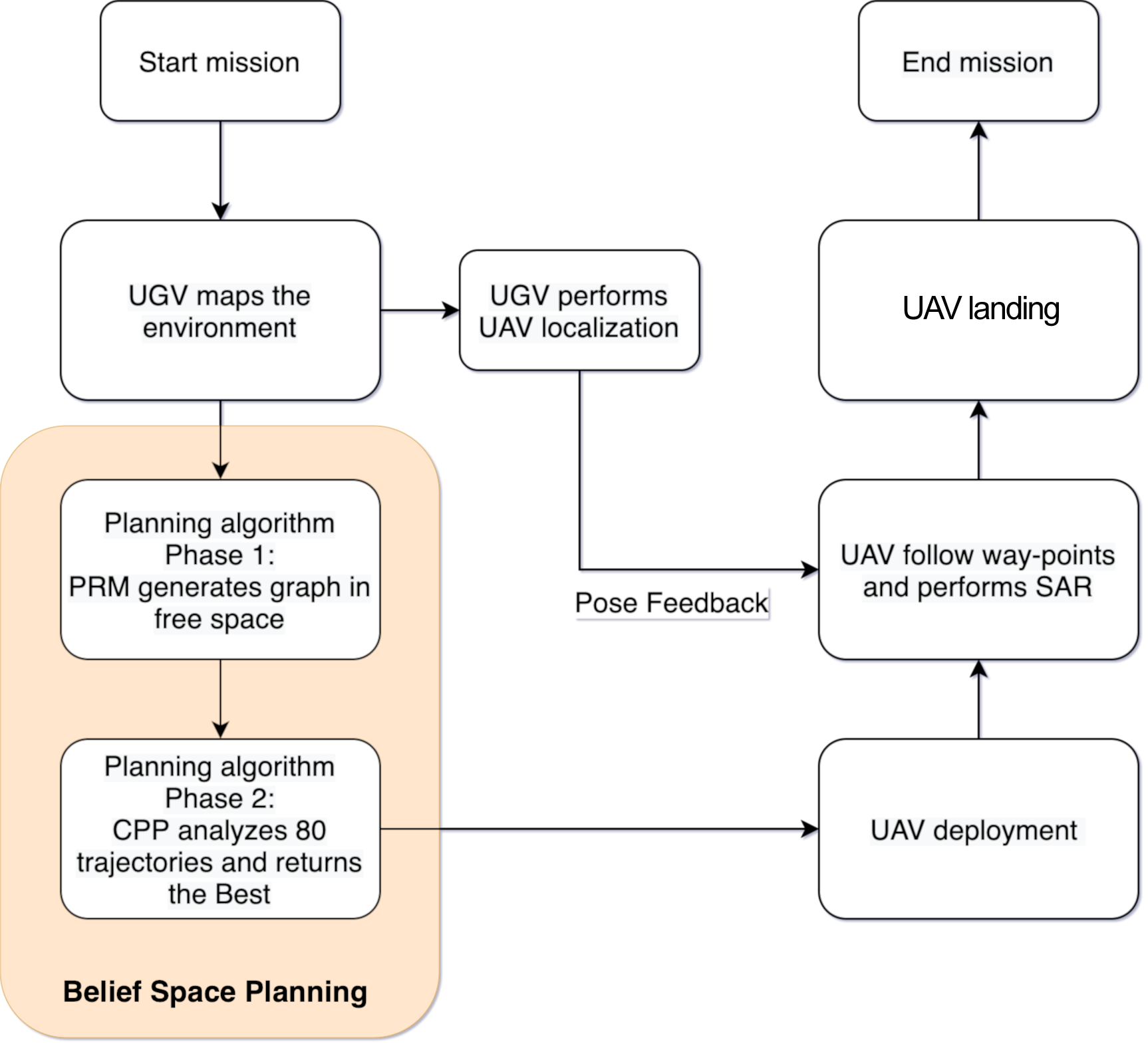}
    \caption{System block diagram representation}
    \label{diagram}
\end{figure}

The algorithm consists of two phases and decouples the exploration task from the criterion to reduce the localization uncertainty. First, using the the construction phase of the Probabilistic Road Map (PRM) algorithm \cite{kavraki1998analysis}, a graph is built on 12 randomly generated nodes lying in the UAV collision-free space favoring the map in front of the UGV where its LIDAR has maximum coverage. The PRM connects each node using a 5-nearest-neighbour approach selecting only the one in the collision-free space. This leads to graph with 12 nodes and 32 edges, G(N,E) = G(12,32). Then, with the edges of this graph, the Route Inspection Problem (or Chinese Postman Problem) \cite{cpp} is solved. Next, in the second phase, for each of the possible trajectories that meet the exploration criterion, the belief-state is propagated and used to select a path that has suitable position error uncertainty. The overall approach is shown in Figure \ref{diagram}.

\subsection{Paths Generation}
In the first step, the planner generates a connected graph with randomly selected nodes that correspond to possible 3D waypoints for the UAV. Each node that lies in the free space is connected to its 5-nearest nodes and an edge between them is generated if it does not cross an obstacle. To connect the \textit{source} to the \textit{goal}, a path is generated following the Chinese Postman Problem (CPP) \cite{cpp}. In the case where a graph is Eulerian, meaning those where all nodes have an even number of incident edges, the CPP solution involves solving an Eulerian circuit. Eulerian circuits are defined as a trail utilizing every edge of a graph exactly once. 

While there are solvers for non-Eulerian graphs, this scenario permits modification of the graph. As such, it was decided to exploit this to produce an Eulerian graph by connecting nonadjacent odd nodes where permissible, to make the CPP in the true sense, having no reused edges. Because the coverage was accounted for in the density of the graph edges, and belief considerations in the query phase, the CPP solver needed only consider the circuit length. An interesting feature of Eulerian circuits when edge lengths are fixed is the equivalence of costs for all solutions, meaning a random solver would be sufficient. From a designated initial position, the algorithm randomly appends adjacent vertices until the initial node is exhausted of edges. Then, recursing, the planner selects a vertex with unused edges to use as a starting position in the induced graph. This new trail replaces the vertex at one instance in the original trail. This is repeated until all edges are exhausted.

To increase the variety of the solutions while keeping the same nodes, the CPP solver is run 80 times so that for each run all edges are always visited once but in a different order. The number of runs has been chosen for computation purposes related to the CPU capacity of the hosting computer.
\\
\subsection{Belief State Propagation and Extended Kalman Filter Description}

In this section, for the sake of completeness, the details of the error-state Extended Kalman Filter (EKF) implementation are provided and more details of nonlinear estimation can be found in \cite{kalman2}. The EKF's error covariance updates are used to propagate the belief state for planning and consequently, the entire EKF implementation is used to evaluate the estimation performance in a simulated environment. In particular, for each potential path provided by the graph construction phase, the UAV's estimated error covariance is propagated over the entire path taking into account all the sensors' models and their associated update rates along the path. This EKF implementation is very similar to our prior work that is reported in \cite{gross2019field}, however, herein it has been modified such that a kinematic model is used to propagate UAV error covariance between measurement updates, and no ``hover constraints", this low velocity measurement update comes when the UAV is hovering, are used to limit velocity drift. 

Once the belief state of the paths generated from the query phase is conducted, the algorithm selects the path with the lowest total norm of the position error covariance estimates. 

In this implementation, the UAV state vector consists of the UAV velocity and position in a local North-East-Down, navigation frame, whose origin is centered at the UGV position.
\begin{equation}
    \label{state}
    \hat{x}=\lbrack   v_N,  v_E,  v_D,  r_N,  r_E,  r_D \rbrack^T
\end{equation}
The error-state of the Extended Kalman Filter, estimates small perturbations ($\delta$) from the unknown true state vector \eqref{state}  as listed in Eq. \eqref{error-state},
\begin{equation}
\begin{split}
    \label{error-state}
    \delta{ \hat {x}}=\lbrack &  \delta v_N,  \delta v_E,  \delta v_D,  \delta r_N,  \delta r_E,  \delta r_D  \rbrack^T
    \end{split}
\end{equation}
For the altimeter, an absolute distance to the ground is measured from a sensor that is mounted in fixed orientation on the UAV's body. This yields a measurement model that depends upon the UAV's altitude.
\begin{equation}
    \label{alt}
    \hat{z}_{alt} = \frac{- \hat{r}_D}{cos(\theta)cos(\phi)} + v_{alt},
\end{equation}
For this measurement, the covariance matrix is assumed to be $R_{alt} = \sigma_{alt}^2 = 0.01$ $[m^2]$ and the Jacobian of the observation model is given as:
\begin{equation}
H_{alt} = \begin{bmatrix}
        \mathbf{0_{1,5}}&-\frac{1}{cos(\theta)cos(\phi)}
    \end{bmatrix}.
    \label{halt}
\end{equation}\\
In Eq. \eqref{halt} the UAV's pitch $\phi$ and roll $\theta$ are estimated from the UAV's IMU during the belief-state propagation process. The resulting altimeter error covariance matrix update $P_{k|k}$ is given as
\begin{equation}
\begin{split}
    P_{k|k}=\left(I_{6}-KH_{alt} \right) &P_{k|k-1}^{-1} \left(I_{6}-KH_{alt} \right)^T+\\&+ KR_{alt}K^T
    \label{Pupalt}
    \end{split}
\end{equation}
where $K=\left(P_{k|k-1}H\right)^T \left(R_{alt} + H_{alt}P_{k|k-1}H_{alt}^T \right)^{-1}$ is the Kalman gain and $I_{6}$ is the identity matrix with the subscript dimension. Consequently the error-state update is
\begin{equation}
    \hat{\delta x}_{_{k | k}} = \hat{\delta x}_{_{k | k-1}} + K(\hat{z}_{alt} - \hat{r}_{D}).
    \label{dxup}
\end{equation}

The UWB ranging radio measurement is modeled as 
\begin{equation}
    \label{range}
    \hat{z}_{UWB}=\hat{d} +v_{uwb}
\end{equation}
with the following observation model
\begingroup
\begin{equation}
H_{uwb} = \begin{bmatrix}
        \mathbf{0_{1,3}}&-\frac{\hat{r}_{N_{k|k-1}}}{\hat{d}}&-\frac{\hat{r}_{E_{k|k-1}}}{\hat{d}}&-\frac{\hat{r}_{D_{k|k-1}}}{\hat{d}}
    \end{bmatrix}
    \label{huwb}
    \end{equation}
where  $d$ is the estimated distance between the UAV and UGV.
 \begin{equation}
     \hat{d} = \left\Vert
     \begin{bmatrix} \hat{r}_{N_{k|k-1}} &\hat{r}_{E_{k|k-1}} &\hat{r}_{D_{k|k-1}} 
     \end{bmatrix} \right\Vert_2
     \label{dist}
 \end{equation}
This yields the following error-covariance update.
\begin{equation}
\begin{split}
    P_{k|k}=\left(I_{6}-KH_{uwb} \right) &P_{k|k-1}^{-1} \left(I_{6}-KH_{uwb} \right)^T+\\&+ KR_{uwb}K^T
    \label{PupUWB}
    \end{split}
\end{equation}
 
 For the state transition model, under the assumptions of constant velocity dynamics, the state transition matrix $\Phi$ is modeled, using simple kinematics, as
\begingroup
 \begin{equation}
     \Phi = 
     \begin{bmatrix}
     
     \mathbf{I_{3}}&\mathbf{0_{3,3}}\\
     \mathbf{I_{3}}\cdot T_s&\mathbf{I_{3}}
     \end{bmatrix}
 \end{equation}
 \endgroup
 and the error covariance matrix update as
  \begin{equation}
   P_{k|k} = \Phi P_{k|k-1}\Phi^T + Q
 \end{equation}
where $Q$ is the assumed covariance matrix for process noise and $T_s$ is the update rate (sampling time) of $50Hz$.
 \begin{equation}
     Q =  \begin{bmatrix} 
    0.01 &0	    &0	&0	&0  &0\\
    0   &0.01	&0	&0	&0	&0\\
    0   &0	&0.01 &0	&0	&0\\
    0   &0	&0	&1	&0	&0\\
    0   &0	&0	&0	&1	&0\\
    0   &0	&0	&0	&0	&1
     \end{bmatrix}
     \label{q}
 \end{equation}

While each of the previous updates occurs at scheduled intervals or update rates, the LIDAR and the fish-eye camera measurements occur only when the UAV is in their respective fields-of-view. As detailed in \cite{gross2019field}, the fish-eye camera position estimate of the UAV assumes no significant motion of the UGV between frames in order to use a simple background subtraction method from the OpenCV library to distinguish the UAV from the surroundings \cite{opencv}.  The measurement update provides an estimate of the unit-vector that points to the UAV with respect to the UGV 
\begin{equation}
    \label{losObs}
    \hat{z}_{cam}=\begin{bmatrix} \frac{\hat{r}_N}{d} \\
    \frac{\hat{r}_E}{d} \\
    \frac{\hat{r}_D}{d} 
\end{bmatrix} +v_{3\times1, cam}
\end{equation}
and the camera update's observation model is given as,

  \begin{equation}
        H_{cam} = \begin{bmatrix}
            \mathbf{0_{1,3}} 
            &\tfrac{-(\hat{r}_E^2+\hat{r}_D^2)}{{\hat{d}}^3} 
            &\tfrac{\hat{r}_N\hat{r}_E}{{\hat{d}}^3}
            &\tfrac{\hat{r}_N\hat{r}_D}{{\hat{d}}^3}
            \\[6pt]
            \mathbf{0_{1,3}}
            &\tfrac{\hat{r}_N\hat{r}_E}{{\hat{d}}^3}
            &\tfrac{-(\hat{r}_N^2+\hat{r}_D^2)}{{\hat{d}}^3}
            &\tfrac{\hat{r}_E\hat{r}_D}{{\hat{d}}^3}
            \\[6pt]
            \mathbf{0_{1,3}}
            &\tfrac{\hat{r}_N\hat{r}_D}{{\hat{d}}^3}
            &\tfrac{\hat{r}_E\hat{r}_D}{{\hat{d}}^3}
            &\tfrac{-(\hat{r}_N^2+\hat{r}_E^2)}{{\hat{d}}^3}
        \end{bmatrix}
    \end{equation}

 where $d$ is as Eq. \eqref{dist} and the covariance matrix update of the form
\begin{equation}
\begin{split}
  P_{k|k}=\left(I_{6}-KH_{cam} \right) P_{k|k-1}^{-1} &\left(I_{6}-KH_{cam} \right)^T +\\&+ K(\delta R_{cam})K^T
   \label{Pcam}
   \end{split}
 \end{equation}
 with the scale factor $\delta = 1/sin(\alpha)$, where $\alpha = \tan^{-1}(-\frac{z}{||z||})$.\\
 
 Finally, the update from the LIDAR sensor provides an observation of the UAV's position within the LIDAR point-cloud as explained in detail in \cite{gross2019field}. In particular, once the point-cloud is segmented into a most likely volume using the UWB and altimeter data, a fast depth-image clustering technique ~\cite{bogoslavskyi2017efficient} is adopted to detect the UAV within a bounding box.  The resulting measurement model assumes that the UAV's position is located at the center of the estimated bounding box.
 \begin{equation}
    \label{NED}
    \hat{z}_{LIDAR} = \begin{bmatrix}
    \hat{r}_N \\
    \hat{r}_E \\
    \hat{r}_D
    \end{bmatrix} +v_{3\times1,{LIDAR}}
\end{equation} 
 This yields an observation model matrix of
  \begin{equation}
        H_{_{{LIDAR}}} = \begin{bmatrix}
            \mathbf{0_{3,3}}&\mathbf{-I_{3}}
        \end{bmatrix},
    \end{equation}
 and covariance matrix update of
\begin{equation}
\begin{split}
    P_{k|k}=\left(I_{6}-KH_{_{{LIDAR}}} \right) P_{k|k-1}^{-1} &\left(I_{6}-KH_{_{{LIDAR}}} \right)^T+\\&+K(\gamma R_{_{{LIDAR}}})K^T
      \end{split}
      \label{PLIDAR}
 \end{equation}
 where $\gamma$ a scale factor that is based up to the number of LIDAR points that are within the determined UAV bounding box. 
 
\section{Simulator Overview}\label{simulator}
To develop and test the planning algorithm for the UAV, a simulator was implemented to closely replicate the physical system detailed in \cite{gross2019field}. For this purpose, the simulator environment was created in ROS (Robot Operating System) \cite{ros} and Gazebo Simulator \cite{Aguero-2015-VRC}. As in the physical system, the PixHawk 4 firmware \cite{px4} was used for the UAV, and it was interfaced with ROS through the MAVROS \cite{mavros} package. This allowed access to the altimeter and many other simulated sensors' data. To simulate the Velodyne LIDAR sensor present on the real UGV, the ROS Velodyne description package was modified to replicate the 128 laser beams available on the rover. Also, an Ultra-Wideband (UWB) ranging radio was simulated to provide measurements of the relative distance between the two UAV and UGV. An upward fish-eye camera was added to the simulator using the wide-angle camera sensor type, to match the real system. The line of sight vector from the UGV to the UAV was then determined with the Scaramuzza D. model \cite{scaramuzza2006flexible}.

Figure \ref{tunnel2} represents the point cloud generated by the Velodyne LIDAR, which is the input to the ROS Octomap \cite{hornung13auro} package to generate a 3D occupancy volume (i.e., an octree) of the environment.
\begin{figure}[h]
    \centering
  \includegraphics[width = 0.95\columnwidth,height=5cm]{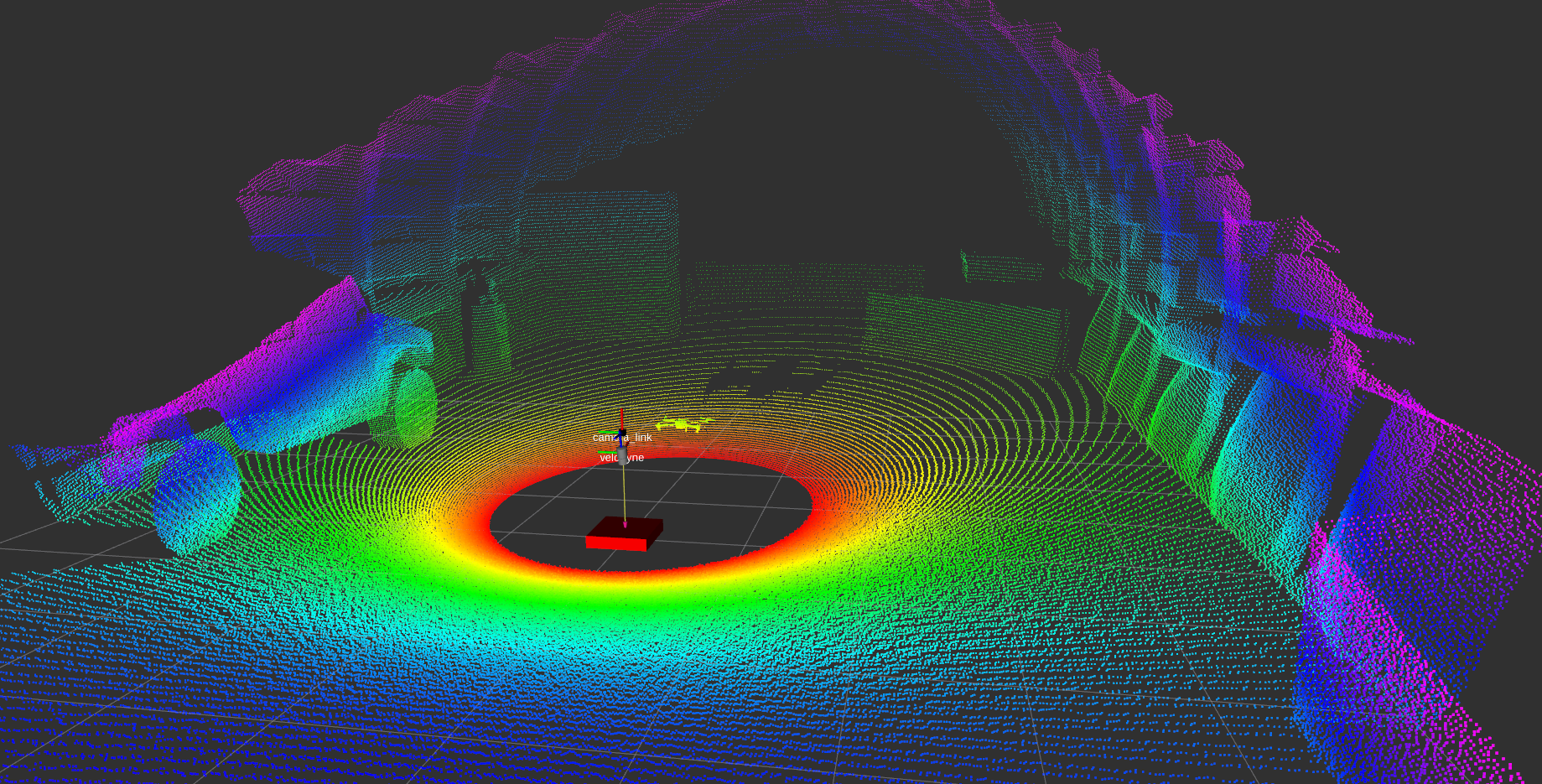}
    \caption{Velodyne LIDAR simulator view}
    \label{tunnel2}
\end{figure}
\begin{figure}[h]
    \centering
     \includegraphics[width = 0.95\columnwidth,height=6cm]{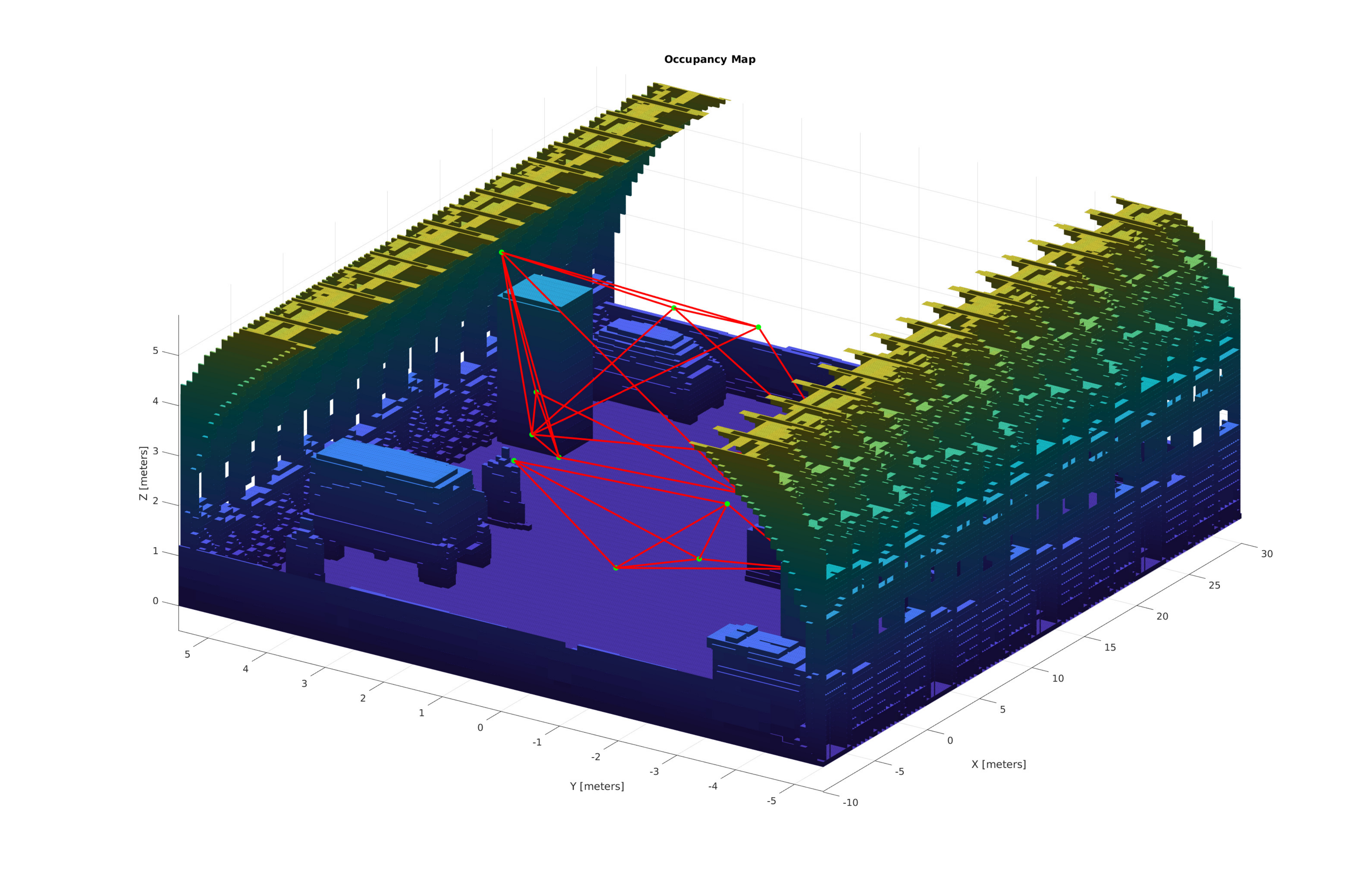}
    \caption{Octomap representation of the environment with the trajectories available}
    \label{tunnel3}
\end{figure}

Using the PixHawk firmware, functions were developed for teleoperation and autonomous flight of the UAV within the simulator. In the autonomous flight mode, the autopilot receives a set of waypoints from the planning approach detailed above and it uses its internal PID controller to reach them.

\section{Results}
\subsection{Path Planning}
The planning algorithm selects waypoints and paths that do not collide with obstacles, generating a graph that lies in the free workspace as shown in Figure \ref{tunnel3} where the 3D occupancy volume of the environment is represented. Next, the planning algorithm was run in order to evaluate 80 different trajectories to find the one that reduces the UAV localization uncertainty. 

Because the first phase of the algorithm visits all edges of the graph in a random order, all trajectories have equivalent costs with respect to the exploration criteria. 
Therefore, for each simulated belief space propagation, first the $\mathcal{L}_2$ norm of the position error covariance is calculated at each time step $i$ as\\
\begin{equation}
PEC_i =  ||P_{pos}^i||.
\label{peci}
\end{equation}
Next, this value was computed over the entire path in order to provide an indication of the quality of the position error covariance of the flight $j$ 
\begin{equation}
    PEC_{flight_j} = \sum_{i=1}^{t} PEC_i
    \label{pecj}
\end{equation}
 where $t$ is the number of time steps in the flight $j$. Finally, the best and worst trajectories were selected by finding the minimum and maximum of these values respectively
 \begin{equation}
     BEST_{path} = \min_{j=1,2,\dots,80}(PEC_{flight_j}),
 \end{equation}
  \begin{equation}
     WORST_{path} = \max_{j=1,2,\dots,80}(PEC_{flight_j}).
 \end{equation}
 Figure \ref{fig:bars} shows the value of each flight as in Eq. \ref{pecj} and in red the worst, green the best, black the second-worst and yellow the second best paths selected by the planning algorithm.
\begin{figure}[!h]
    \centering
    \includegraphics[width = 0.95\columnwidth,height=6cm]{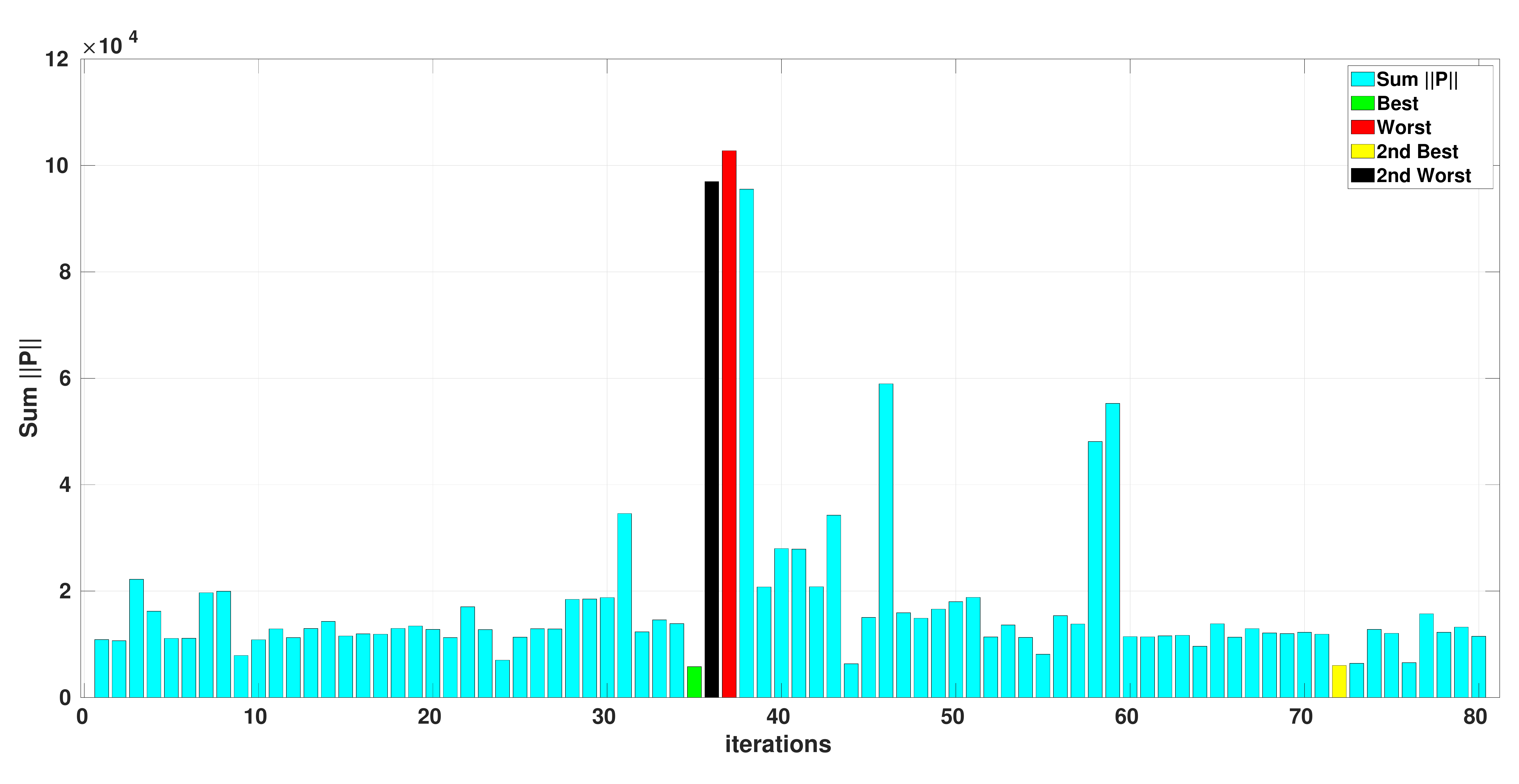}
    \caption{Sum of the $\mathcal{L}_2$ norm of the Position Error Covariance for each trajectory, in red the one with the maximum total position error covariance and in green the one with the minimum. Also in black the path with the second maximum total position error covariance and in yellow the respective second minimum.}
    \label{fig:bars}
\end{figure}
A side-by-side of the estimated uncertainty of the four trajectories selected are also shown for comparison in Table \ref{tab:stats}.

\setlength{\tabcolsep}{13pt} 
\renewcommand{\arraystretch}{1.6} 
\begin{center}
\begin{table}[H]
\centering
\footnotesize{
\caption{ Planning algorithm statistics.}
\label{tab:stats}
\begin{tabular}{m{0.14\linewidth}|m{0.08\linewidth}m{0.08\linewidth}m{0.08\linewidth}m{0.12\linewidth}}
\hline
 & \cellcolor{green!45}\shortstack{\\Best\\Path}  &\cellcolor{red!45}\shortstack{\\Worst\\ Path} &\cellcolor{yellow!45}\shortstack{\\$2^{nd}$ Best\\ Path}  &\cellcolor{gray!45}\shortstack{\\$2^{nd}$ Worst \\Path}\\
\hline
\hline
\vspace{5pt}
\shortstack{3D RMS \\Pos. Err. Cov. \\$(m^2)$}  &120.97     & 2527          &121.13   & 2518.1\\ 
\vspace{5pt}
\shortstack{3D Max. \\Pos. Err. Cov. \\$(m^2)$} & 5796.4    & 1.03e+5     &6026.8   &96947\\ 
% \vspace{5pt}
\shortstack{ $ \sigma (m^2)$}                     & 120.93    & 2526.70       &121.08   & 2518\\
\shortstack{ $ \mu (m^2)$}                        & 4.55   & 80.68        &4.73   & 76.10\\
\shortstack{Median $(m^2)$}                       & 1.0006    & 1.0005        &1.0005    & 1.0005\\
\hline
\hline
% \caption{Important values for the best and worst trajectory}
\end{tabular}
}
\end{table}
\end{center}
\endgroup

\vspace{-13pt}

\subsection{Position Estimation}
The best and worst trajectories selected by the planning algorithm were then simulated in the simulator presented in section \ref{simulator} and compared to show the benefits of the proposed approach. 

It is important to note for this evaluation, when simulating this trajectory, that the UAV's perfectly-known simulated position is used by the flight controller pose estimator as feedback, in order to execute the trajectory. That is, the presented and evaluated EKF's feedback was not used for closed-loop control. This was conducted to facilitate the evaluation of the planner's ability to determine a trajectory without having to consider the implications of poor localization feedback in the execution of the path.  

\begin{figure}[H]
    \centering
 \includegraphics[width =0.95\columnwidth,height=5cm]{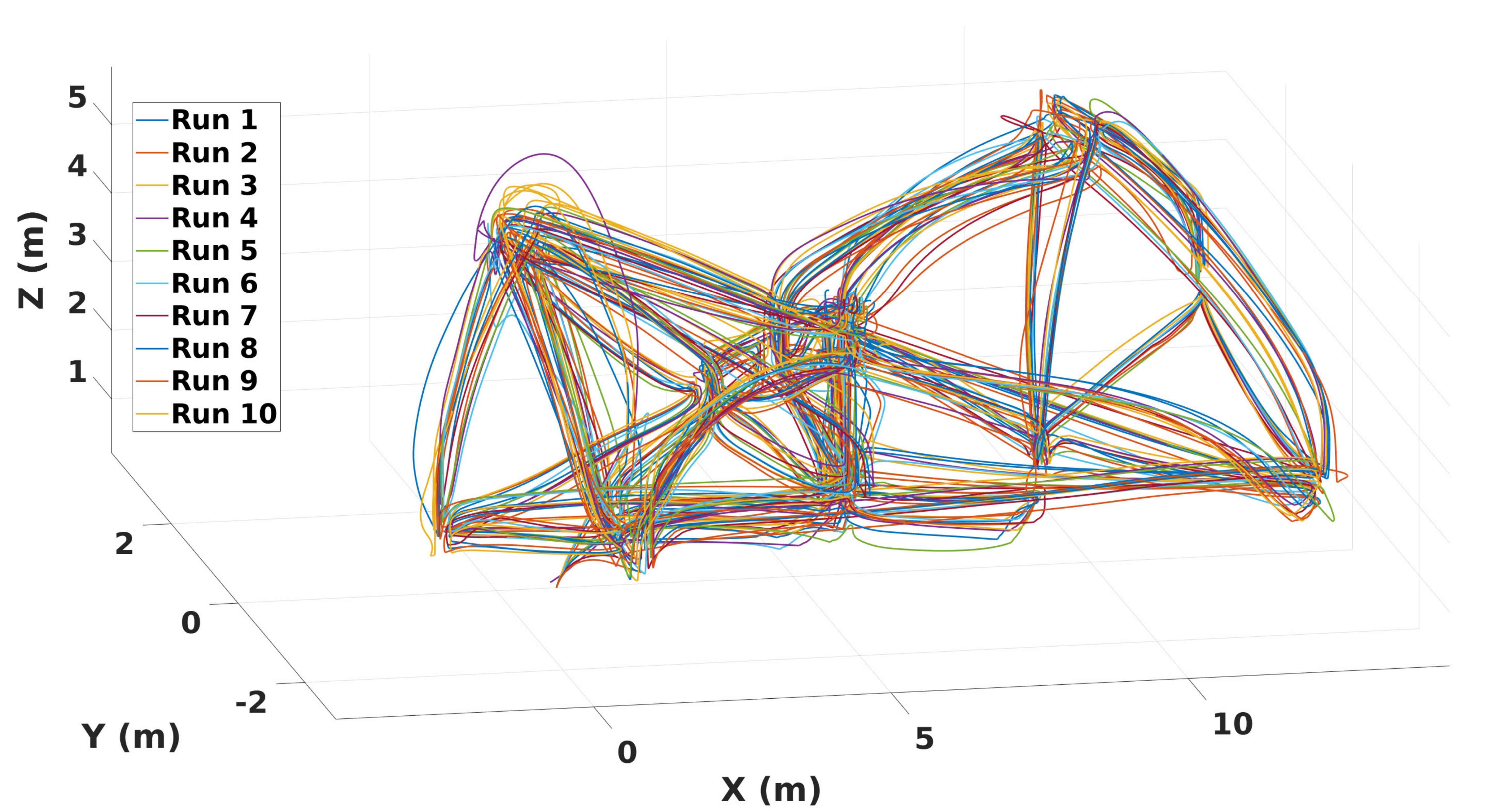}
  \caption{UAV ground truth position of the same set of waypoints run multiple times}
    \label{fig:10truth}
\end{figure}

An ``online" EKF, as detailed above, was then used to estimate the UAV's position with respect to the UGV. In particular, the analysis of ten simulations of each set of waypoints was executed to better understand the trends of the planning algorithm. The use of multiple trials to show trends was conducted because, during the evaluation of the algorithm, it was determined that even when the same list of waypoints was provided to the flight controller in the simulation, the resulting executed paths had variations from trial-to-trial. For example, to show the differences between each simulation that was provided the same set of waypoints, the UAV's ground truth of each run is represented in Figure \ref{fig:10truth}. 

It is expected that this variation is due to the structure of the PX4 firmware, the onboard estimator uses the ground truth as a sensor update to onboard IMU with sensor noise and not as direct pose feedback for the controller loop leading to have slightly different results each time the simulation is run.
In this study, our intention was to focus on the planning algorithm, the UAV's autopilot and estimator have not been changed. For this reason, multiple simulations of the same set of waypoints, that return slightly different results, were conducted such that average trends could be considered. 

\begin{figure}[!h]
    \centering
    \includegraphics[width = 0.95\columnwidth, height=6cm]{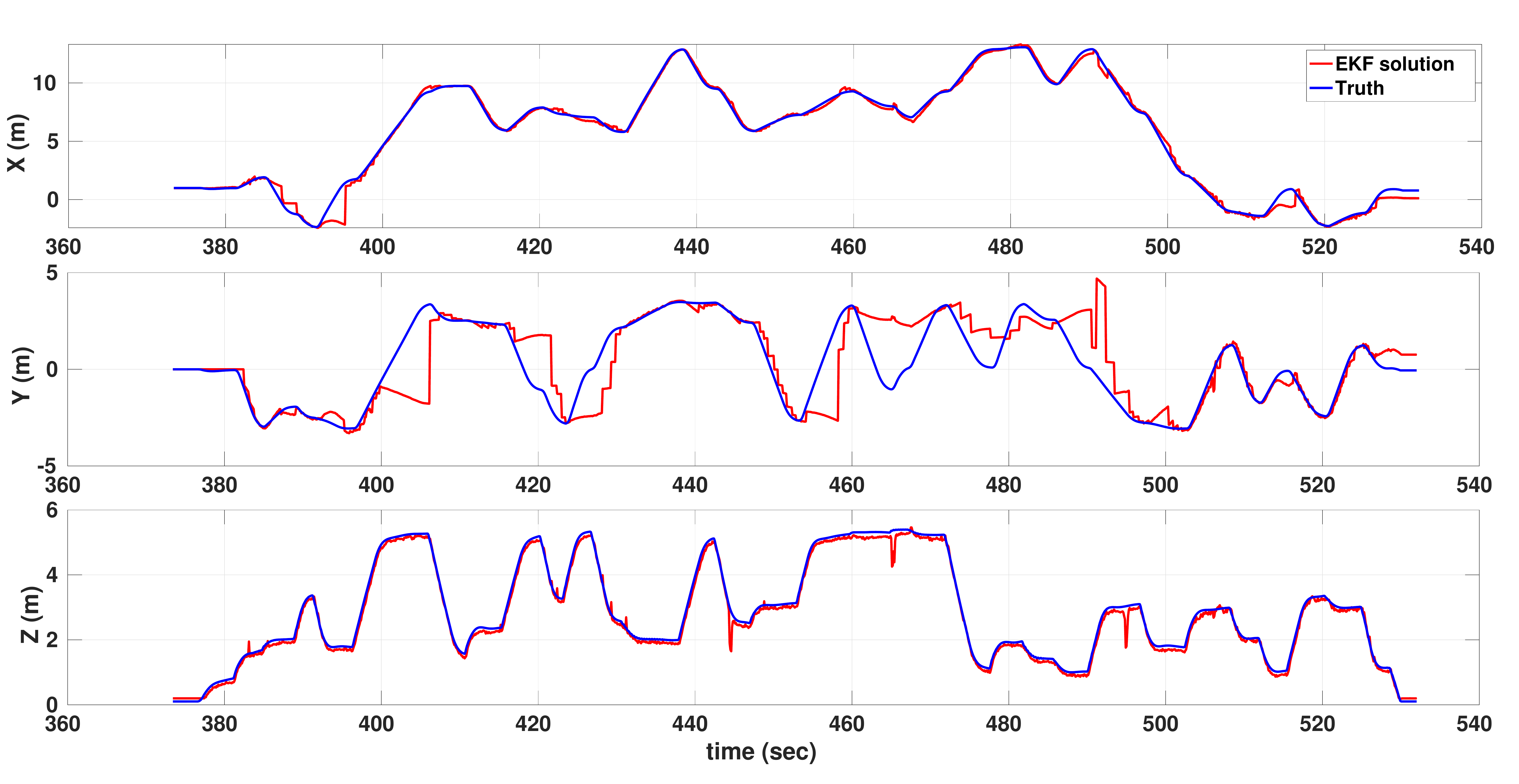}
    \caption{In red the EKF position estimation of the best trajectory, while in blue the ground truth.}
    \label{fig:bestSol}
\end{figure}

\begin{figure}[h]
    \centering
    \includegraphics[width = 0.95\columnwidth,height=6cm]{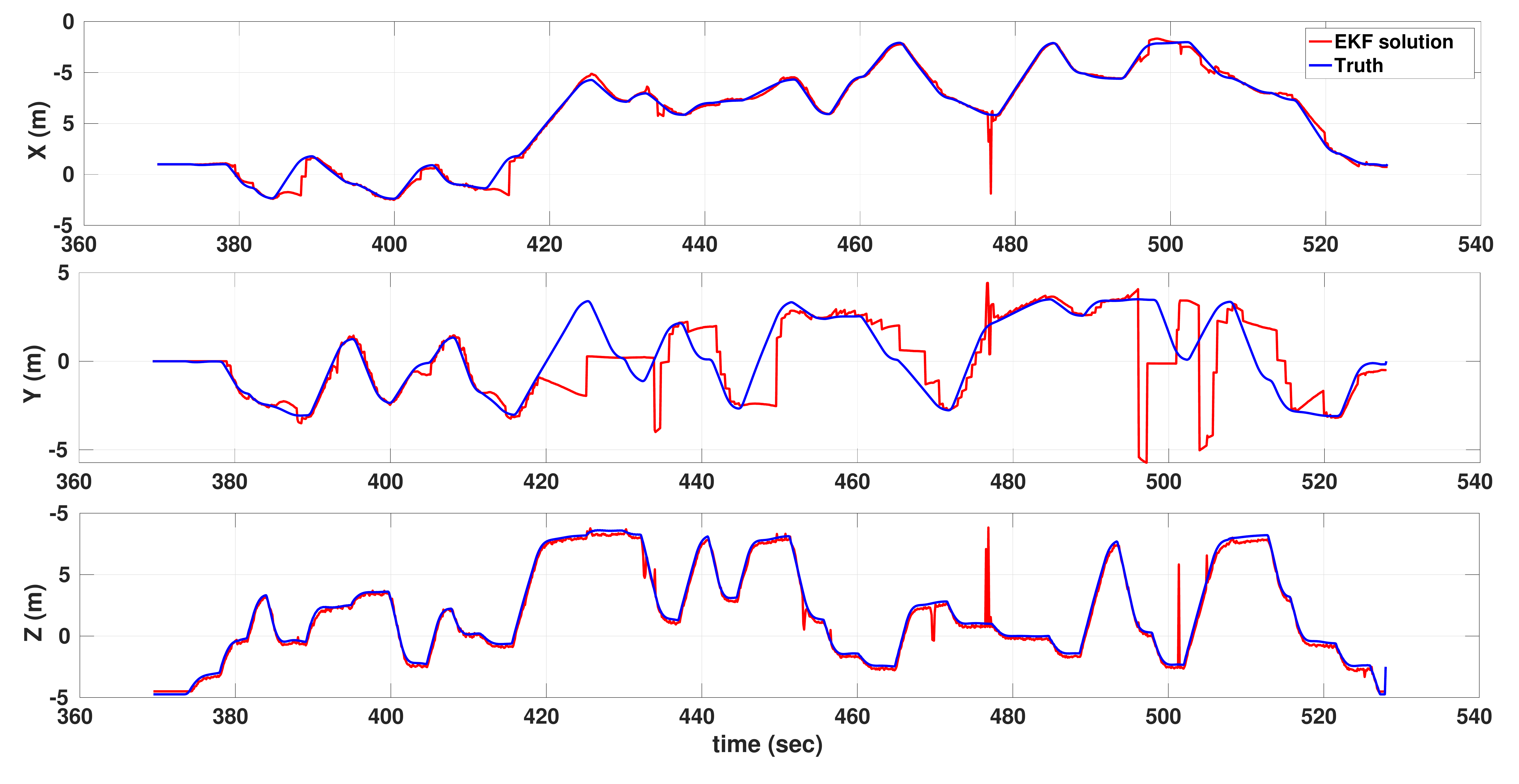}
    \caption{In red the EKF position estimation of the worst trajectory, while in blue the ground truth.}
    \label{fig:worstSol}
\end{figure}

For the ten trials, the UAV's EKF estimated position with the simulation ground truth when executing the best trajectory is shown in Figure \ref{fig:bestSol}. 
In addition, the results of the UAV position estimation for the worst trajectory are presented in Figure \ref{fig:worstSol}, where it is evident when comparing the two cases that EKF performs notably worse. 

The main differences between the best and worst trajectory are summarized in Table \ref{tab:Simstats}, where $\sigma$ and $\mu$ are the standard deviation and mean of the 3D positioning error respectively and almost 0.32 meters 3D RMS improvement between the two paths. The spikes on the Z estimate of the UAV's position, shown in both Figure \ref{fig:bestSol} and Figure \ref{fig:worstSol}, are related to the altimeter's observation model in Eq. \eqref{halt}, which is a function of the IMU estimate of pitch and roll.
\setlength{\tabcolsep}{13pt} 
\renewcommand{\arraystretch}{1.6} 
\begin{center}
\begin{table}[H]
\centering
\footnotesize{
\caption{Comparison between the simulation of the best and worst path.}
\label{tab:Simstats}
\begin{tabular}{c | cc} 
\hline
 &  {Best Path}  & {Worst Path}\\
\hline
\hline
Flight time (s)             & 158.70    & 158.52\\
$X_{RMS}$\hspace{5pt}(m)    & \cellcolor{green!45}0.25     & \cellcolor{red!45}0.54\\
$Y_{RMS}$\hspace{5pt}(m)    & \cellcolor{green!45}1.38     & \cellcolor{red!45}1.68\\
$Z_{RMS}$\hspace{5pt}(m)    & \cellcolor{green!45}0.15     & \cellcolor{red!45}0.19\\
{RMS 3D Pos. Err. (m)}      & \cellcolor{green!45}1.46     &\cellcolor{red!45} 1.77\\
$\sigma$ (m)                & \cellcolor{green!45}1.10     & \cellcolor{red!45}1.42\\
$\mu$ (m)                   & \cellcolor{green!45}0.95     & \cellcolor{red!45}1.07\\
Median (m)                  & \cellcolor{green!45}0.46     & \cellcolor{red!45}0.47\\
Max Pos. (m)                & \cellcolor{green!45}5.24     & \cellcolor{red!45}9.24\\
\# LIDAR Updates            & 91      & 97\\ 
\# Camera Updates           & 162       & 161\\
\hline
\hline
\end{tabular}
}
\end{table}
\end{center}

As listed in Table \ref{tab:Simstats}, for the comparison, the 3D RMS position error is generally smaller for the best trajectory as compared to the worst, reducing the mean of the position error in half. This is also the case with respect to the Maximum Position Error (MPE), mean and standard deviation. 

Due to the variability in performing the selected paths, as shown in Figure \ref{fig:10truth}, the need to present consistent results by the planning algorithm and the navigation filter is essential and in Table \ref{tab:simRuns} are shown the results of ten runs of the best and the worst trajectories.

\setlength{\tabcolsep}{5pt} 
\renewcommand{\arraystretch}{1.4} 
\begin{table}[!h]
\caption{Comparison of multiple runs for the best and worst trajectories.}
\centering
\label{tab:simRuns}
\begin{tabular}{p{0.08\linewidth}|p{0.081\linewidth}p{0.12\linewidth}p{0.1\linewidth}p{0.14\linewidth}p{0.11\linewidth}}
\hline
{Path} &{Run} &{Median} & {Mean }&\shortstack{\\RMS\\(Pos. Err.)} & {MPE}  \\
\hline
\hline
\multirow{10}{1\linewidth}{Best Path}
& 1   & 1.19 &   0.55 &  1.76 &   9.53\\    
& 2   & 1.02 &   0.80 &  1.38 &   5.03\\    
& 3   & 1.15 &   0.52 &  1.90 &  10.09\\    
& 4   & 0.99 &   0.46 &  1.51 &   5.52\\    
& 5   & 1.03 &   0.49 &  1.61 &   7.81\\
& 6   & 0.97 &   0.46 &  1.57 &   7.91\\    
& 7   & 0.95 &   0.46 &  1.46 &   5.24\\    
& 8   & 1.05 &   0.49 &  1.58 &   5.28\\    
& 9   & 1.01 &   0.40 &  1.66 &   6.71\\    
& 10  & 1.01 &   0.50 &  1.54 &   7.36\\ 
\hline
\multirow{10}{1\linewidth}{Worst Path} 
& 1   & 1.16  &  0.43 &  2.16 &  15.68\\
& 2   & 1.54  &  0.54 &  3.56 &  24.63\\
& 3   & 1.22  &  0.53 &  2.08 &   8.19\\
& 4   & 0.92  &  0.49 &  1.37 &   5.33\\
& 5   & 1.07  &  0.47 &  1.77 &   9.24\\
& 6   & 2.12  &  0.56 &  4.55 &  26.23\\
& 7   & 0.98  &  0.44 &  1.60 &   8.21\\
& 8   & 0.98  &  0.43 &  1.58 &   6.62\\
& 9   & 0.90  &  0.40 &  1.42 &   6.39\\
& 10  & 1.86  &  0.70 &  3.64 &  25.12\\
\hline
\hline
\end{tabular}
\end{table}

\setlength{\tabcolsep}{5pt} 
\renewcommand{\arraystretch}{1.65} 
\begin{table}[!h]
\caption{Mean and Median of Table III for best and worst paths.}
\centering
\label{tab:simRunsMean}
\begin{tabular}{p{0.08\linewidth}|p{0.09\linewidth}p{0.12\linewidth}p{0.12\linewidth}p{0.12\linewidth}p{0.12\linewidth}}
\hline
{Path} &{} &{Median} & {Mean }&\shortstack{\\RMS\\(Pos. Err.)} & {MPE}   \\
\hline
\hline
\multirow{2}{1\linewidth}{Best Path}    
& Mean      &	\cellcolor{green!45}1.04&  \cellcolor{red!45}  0.52&   \cellcolor{green!45}1.60&        \cellcolor{green!45} 7.05\\
& Median    &	\cellcolor{green!45}1.02 &  \cellcolor{red!45} 0.49& \cellcolor{green!45}1.57&       \cellcolor{green!45} 7.03\\
\hline
\multirow{2}{1\linewidth}{Worst Path} 
& Mean      &	\cellcolor{red!45}1.27&    \cellcolor{green!45}0.50  & \cellcolor{red!45}2.37 &       \cellcolor{red!45} 13.56\\
& Median    &	\cellcolor{red!45}1.11&    \cellcolor{green!45}0.48   & \cellcolor{red!45}1.93&        \cellcolor{red!45} 8.73\\	
\hline
\hline
\end{tabular}
\end{table}

% --------------------------------------
\setlength{\tabcolsep}{5pt} 
\renewcommand{\arraystretch}{1.4} 
\begin{table}[!h]
\caption{Comparison of multiple runs for the second best and second worst trajectories.}
\centering
\label{tab:simRuns2}
\begin{tabular}{p{0.08\linewidth}|p{0.081\linewidth}p{0.12\linewidth}p{0.12\linewidth}p{0.12\linewidth}p{0.12\linewidth}p{0.12\linewidth}}
\hline
{Path} &{Run} &{Median} & {Mean }&\shortstack{\\RMS\\(Pos. Err.)} & {MPE}   \\
\hline
\hline
\multirow{10}{1\linewidth}{$2^{nd}$\\Best Path}
& 1   & 1.02 &   0.50 &   1.57 &   6.29\\    
& 2   & 1.12 &   0.49 &   1.76 &   6.48\\    
& 3   & 1.48 &   0.61 &   3.12 &  26.96\\    
& 4   & 1.05 &   0.45 &   1.68 &   6.47\\    
& 5   & 1.27 &   0.58 &   2.03 &   8.20\\    
& 6   & 1.05 &   0.49 &   1.67 &   7.47\\    
& 7   & 1.20 &   0.52 &   1.84 &   6.87\\    
& 8   & 1.11 &   0.61 &   1.68 &   7.44\\    
& 9   & 0.98 &   0.46 &   1.55 &   6.83\\    
& 10  & 1.09 &   0.51 &   1.66 &   7.74\\ 
\hline
\multirow{10}{1\linewidth}{$2^{nd}$\\Worst Path} 
& 1   & 1.32 &   0.47 &  2.32 &   9.56\\    
& 2   & 0.97 &   0.42 &  1.52 &   6.53\\    
& 3   & 1.00 &   0.43 &  1.68 &   8.09\\    
& 4   & 1.13 &   0.54 &  1.75 &   8.43\\    
& 5   & 0.90 &   0.42 &  1.44 &   6.08\\    
& 6   & 1.01 &   0.53 &  1.47 &   5.61\\    
& 7   & 1.07 &   0.53 &  1.59 &   5.79\\    
& 8   & 0.98 &   0.41 &  1.62 &   7.19\\    
& 9   & 1.08 &   0.47 &  1.67 &   6.19\\    
& 10  & 2.75 &   0.59 &  5.89 &  25.06\\    
\hline
\hline
\end{tabular}
\end{table}

\setlength{\tabcolsep}{5pt} 
\renewcommand{\arraystretch}{1.65}
\begin{table}[!h]
\caption{Mean and Median of Table V for the $2^{nd}$ best and $2^{nd}$ worst paths.}
\centering
\label{tab:simRunsMean2}
\begin{tabular}{p{0.08\linewidth}|p{0.09\linewidth}p{0.12\linewidth}p{0.12\linewidth}p{0.12\linewidth}p{0.12\linewidth}}
\hline
{Path} &{} &{Median} & {Mean }&\shortstack{\\RMS\\(Pos. Err.)} & {MPE}  \\
\hline
\hline
\multirow{2}{1\linewidth}{$2^{nd}$\\Best Path}    
& Mean    &	\cellcolor{green!45}1.14  & \cellcolor{red!45} 0.52& \cellcolor{green!45}1.86 &   \cellcolor{red!45}    9.08\\
& Median  &	\cellcolor{red!45}1.10&  \cellcolor{red!45}   0.50& \cellcolor{red!45}1.68&     \cellcolor{red!45}   7.16\\
\hline
\multirow{2}{1\linewidth}{$2^{nd}$\\Worst Path} 
& Mean   &\cellcolor{red!45}1.22   & \cellcolor{green!45}0.48 &\cellcolor{red!45}	 2.09&        \cellcolor{green!45}8.85 \\
& Median &	\cellcolor{green!45} 1.04 &   \cellcolor{green!45}0.47& \cellcolor{green!45}1.65  &      \cellcolor{green!45}6.86\\
\hline
\hline
\end{tabular}
\end{table}

To easily understand the data in Table \ref{tab:simRuns} the mean and median of those values are available in Table \ref{tab:simRunsMean}. From Table \ref{tab:simRuns} can be shown that not always the use of the best trajectory will localize the UAV better in absolute value, while with multiple runs it is demonstrated that the planning approach is beneficial for reducing UAV position error.

To demonstrate the repeatability of the planning algorithm, an additional second set of simulated trajectories is presented in Table \ref{tab:simRuns2} and Table \ref{tab:simRunsMean2} where now the ``second" best and ``second" worst trajectories were considered. As shown in Table \ref{tab:simRuns2} the second best path does not perform as expected providing often worse results than the second worst path. This is further described in the next section, with our belief of the most likely explanations.

\subsection{Belief Space Planning vs Simulation}
To better demonstrate the effectiveness of the planning algorithm and offer insight as to why we have seen some inconsistencies in results obtained in Table \ref{tab:simRuns2}, additional tests have been conducted using a ``perfect'' version of the navigation filter that is free from the possibility of missing or outlier measurement updates of the UAV position from the UGV's camera and LIDAR system. In particular, in this scenario, whenever the UAV was within the simulated modeled field-of-view of the camera or LIDAR, the truth position of the UAV was used to construct measurement updates for the EKF. This was to remove the potential impact of missing sensors' updated during the simulation that were caused by detection issues (e.g., not seeing the UAV or incorrectly clustering something as the UAV) and were therefore not reflected in the planning algorithm.  
\\Similarly to the previous section, using this ``perfect'' EKF, ten runs of the four paths (best, worst, $2^{nd}$ best, $2^{nd}$ worst) were run and the mean and median are reported in Tables \ref{tab:p1} and \ref{tab:p2}.

\setlength{\tabcolsep}{5pt} 
\renewcommand{\arraystretch}{1.65} 
\begin{table}[!h]
\caption{Mean and Median of 10 runs of the best and the worst trajectories assuming perfect measurements.}
\centering
\label{tab:p1}
\begin{tabular}{p{0.08\linewidth}|p{0.09\linewidth}p{0.12\linewidth}p{0.12\linewidth}p{0.12\linewidth}p{0.12\linewidth}}
\hline
{Path} &{} &{Median} & {Mean }&\shortstack{\\RMS\\(Pos. Err.)} & {MPE}  \\
\hline
\hline
\multirow{2}{1\linewidth}{Best Path}    
& Mean    &	\cellcolor{green!45}0.40  & \cellcolor{green!45} 0.57& \cellcolor{green!45}0.81 &   \cellcolor{green!45}    4.22\\
& Median  &		\cellcolor{green!45}0.41  & \cellcolor{green!45} 0.57& \cellcolor{green!45}0.80 &   \cellcolor{green!45}    4.39\\
\hline
\multirow{2}{1\linewidth}{Worst Path} 
& Mean   &\cellcolor{red!45} 0.42  & \cellcolor{red!45}0.61 &\cellcolor{red!45}	 0.90&        \cellcolor{red!45}5.12 \\
& Median &	\cellcolor{red!45} 0.43 &   \cellcolor{red!45}0.61& \cellcolor{red!45}0.89  &      \cellcolor{red!45}4.47\\
\hline
\hline
\end{tabular}
\end{table}

\setlength{\tabcolsep}{5pt} 
\renewcommand{\arraystretch}{1.65} 
\begin{table}[!h]
\caption{Mean and Median of 10 runs of the second best and second worst trajectories assuming perfect measurements.}
\centering
\label{tab:p2}
\begin{tabular}{p{0.08\linewidth}|p{0.09\linewidth}p{0.12\linewidth}p{0.12\linewidth}p{0.12\linewidth}p{0.12\linewidth}}
\hline
{Path} &{} &{Median} & {Mean }&\shortstack{\\RMS\\(Pos. Err.)} & {MPE}  \\
\hline
\hline
\multirow{2}{1\linewidth}{$2^{nd}$\\Best Path}    
& Mean    &	\cellcolor{green!45}0.40  & \cellcolor{green!45} 0.62& \cellcolor{green!45}0.95 &   \cellcolor{red!45}    5.60\\
& Median  &	\cellcolor{green!45}0.41&  \cellcolor{green!45}   0.62& \cellcolor{green!45}0.94&     \cellcolor{red!45}   5.40\\
\hline
\multirow{2}{1\linewidth}{$2^{nd}$\\Worst Path} 
& Mean   &\cellcolor{red!45} 0.42  & \cellcolor{red!45}0.63 &\cellcolor{red!45}	 0.96&        \cellcolor{green!45}5.32 \\
& Median &	\cellcolor{red!45} 0.42 &   \cellcolor{red!45}0.63& \cellcolor{red!45}0.96  &      \cellcolor{green!45}5.24\\
\hline
\hline
\end{tabular}
\end{table}
As shown in the previous results, when using a navigation filter with no missing or outlier measurement updates and, therefore, more closely reflects the belief space planning approach, the algorithm provides a better choice of waypoints to minimize the overall UAV's localization. These results are meant to illustrate that discrepancies that exist between the belief offline propagation model and the simulation scenario in real-time compromises the reliability of this planning algorithm, which may be the case for the results shown in Table \ref{tab:simRuns2}. 

\section{Conclusions}
With this paper, an offline planning algorithm to find a path in the 3D space that meets search criteria and considers position uncertainty is discussed. In a simulation environment, the approach was shown to offer promise for selecting the order of pre-defined waypoints in order to reduce the UAV's position uncertainty. Extensive simulations have been conducted to demonstrate the benefits of the algorithm. In particular, we showed the capability of the planning algorithm to choose the path where the UAV is more favorable with respect to the discussed EKF estimator. 

Future work will focus on improving the planning algorithm to take into account the issues related to having missing expected measurement updates and/or  outlier measurement updates, as these discrepancies diminish the value of the offline belief space plan. Determining a means to augment the plan online to account for this, provide more consistent results. Future steps also consists of integrating the algorithm on a physical system in order to plan and fly autonomously using in closed-loop the estimated position. In addition, to further reduce localization uncertainty, we plan to include visual-inertial odometry within the EKF. Additional refinements will be made to allow the UAV and UGV to move at the same time to increase exploration.

\bibliographystyle{IEEEtran}
\bibliography{IEEE-AES-System-Magazine-manuscript}

\end{document}